\pdfoutput=1

\documentclass[11pt]{article}

\usepackage{etoolbox}

\usepackage[final]{acl}

\usepackage{times}
\usepackage{latexsym}

\usepackage[T1]{fontenc}

\usepackage[utf8]{inputenc}

\usepackage{microtype}

\usepackage{inconsolata}

\usepackage{graphicx}
\usepackage{booktabs} 

\usepackage{todonotes}
\usepackage{multirow} 

\title{Converting Annotated Clinical Cases into Structured Case Report Forms}

\author{
  \textbf{Pietro Ferrazzi\textsuperscript{1,2}},
  \textbf{Alberto Lavelli\textsuperscript{1}},
  \textbf{Bernardo Magnini\textsuperscript{1}},
  \\
   \textsuperscript{1}Fondazione Bruno Kessler, Via Sommarive 18, Povo, Trento, Italy,
   \\
   \textsuperscript{2} University of Padova,
via Trieste 63, Padova, Italy
  \\
  \small{
    \textbf{Correspondence:} \href{mailto:email@domain}{pferrazzi@fbk.eu}
  }
}

\begin{document}
\maketitle
\begin{abstract}
Case Report Forms (CRFs) are largely used in medical research as they ensure accuracy, reliability, 
and validity of results in clinical studies. However, publicly available, well-annotated CRF datasets are scarce, limiting the development of CRF slot filling systems able to fill in a CRF from clinical notes. To mitigate the scarcity of CRF datasets, we propose to take advantage of available datasets annotated
for information extraction tasks
and to convert them into structured CRFs. We present a semi-automatic conversion methodology, which has been applied to the E3C dataset in two languages (English and Italian), resulting in a new, high-quality dataset for CRF slot filling. Through several experiments on the created dataset, we report that slot filling achieves 59.7\% for Italian and 67.3\% for English on a closed 
Large Language Models
(zero-shot) and worse performances on three families of open-source models, showing that filling CRFs is challenging even for recent state-of-the-art LLMs.
\end{abstract}

\section{Introduction}
Case Report Forms (CRFs) are essential tools in clinical research, designed to systematically and consistently collect patient data. They are composed of a list of predefined items to be filled with patients' medical information. By standardizing data collection, they ensure accuracy, reliability, and validity, which are crucial for producing meaningful and reproducible results in clinical studies.\\
An expanding area of research focuses on developing automated systems for filling CRFs with information extracted from clinical notes and medical records, a concept envisioned by \citet{MacKenzie2016} and further advanced by \citet{Gutierrez-Sacristan2024}.
Leveraging Natural Language Processing methods and models represent a potentially promising approach to automate and advance research in this field. However, despite their importance, publicly available, well-annotated CRF datasets are scarce, limiting the effective development and training of such systems.

\begin{figure}[t]
\centering
\includegraphics[width=1\linewidth]{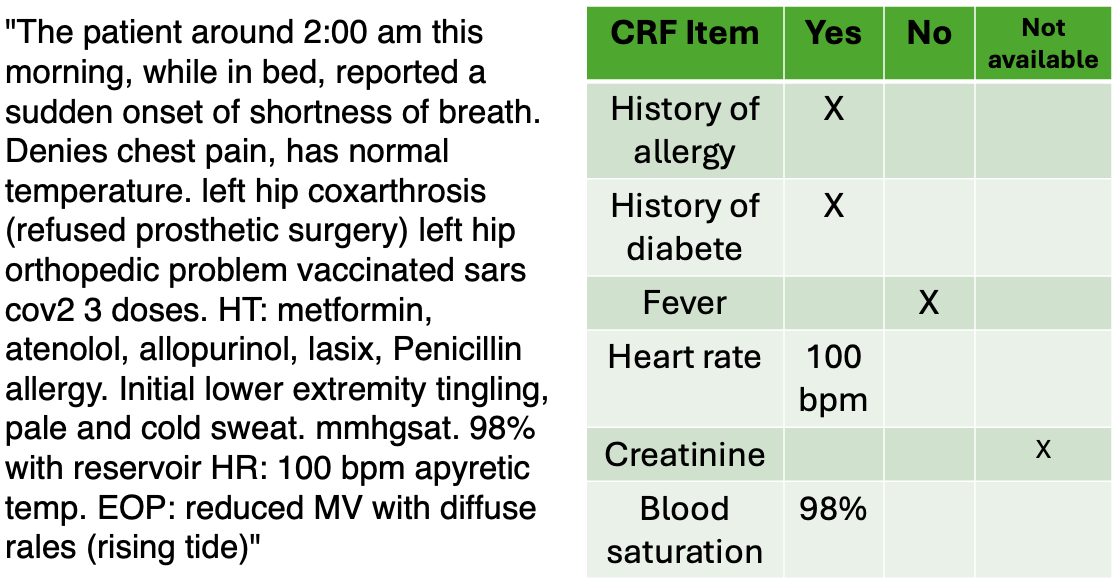}
\caption{Example of a Case Report Form filled with the values from a clinical note.
}
\label{fig:example_note_and_crf}       
\end{figure}

To address this gap, we propose a methodology that transforms publicly available datasets of clinical cases annotated for information extraction into a structured set of filled CRFs.
Examples of such publicly available datasets are the following: MIMIC IV\footnote{\url{https://physionet.org/content/mimiciv/3.1/}}, i2b2\footnote{\url{https://www.i2b2.org/NLP/DataSets/Main.php}}, n2c2\footnote{\url{https://n2c2.dbmi.hms.harvard.edu/}}, CAS~\citep{grabar-etal-2018-cas}, E3C~\citep{Magnini2023}. 
Our approach reduces the discrepancy between existing datasets and real-world clinical needs, aligning them more closely with the practical requirements of hospitals and clinical research applications, where CRF filling is a widely relevant task.
The outcome is a diverse CRF dataset, filled with information grounded in human annotations. Each example in the dataset consists of a triplet: a clinical case, a CRF to be filled, and the golden-standard filling values for the CRF derived from the clinical note, similar to what is presented in Figure~\ref{fig:example_note_and_crf}.
We apply this methodology to the European Clinical Case Corpus (E3C), release the resulting dataset, and evaluate several Large Language Models (LLMs) on it.\\
The contributions of the paper are the following: (i) a general methodology for converting corpora of clinical cases annotated for information extraction into filled CRFs; (ii) a new multilingual dataset\footnote{The dataset is released at \url{https://huggingface.co/collections/NLP-FBK/e3c-to-crf-67b9844065460cbe42f80166}} (Italian and English) for CRF slot filling derived from the E3C dataset; (iii) several baselines indicating that automatic CRF slot filling from clinical notes is challenging even for state-of-the-art LLMs.

\begin{figure*}[t]
\centering
\includegraphics[width=1\linewidth]{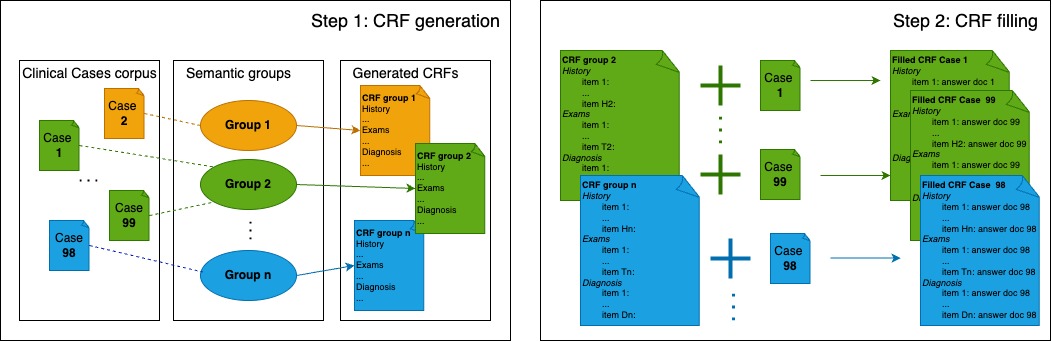}
\caption{Summary of our two-step CRF generation and filling pipeline. \textbf{Step 1} Initially, clinical notes are clustered based on semantic similarity. Then, a group-specific CRF is generated for each cluster by extracting relevant items from the annotations of the clinical cases within the group. \textbf{Step 2} Each case is then linked to its designated CRF, which item set is populated based on the preexisting document annotation. The outcome is a list of as many CRFs as identified groups, and as many filled CRFs as documents. Each group-specific CRF is filled as many times as the number of documents belonging to it.  
}
\label{fig:pipeline}       
\end{figure*}

\section{Related Work}

Health data standardization is a fundamental aspect in the ongoing integration of medical research and artificial intelligence. To facilitate such alliance, the dimensions emphasized by \citet{Petavy2019} are crucial, encompassing the need of health research for being transparent, accessible, interoperable, reproducible, and of high quality.\\
Case Report Forms play a central role in this context, and various efforts have been made to ensure that CRFs are designed to be consistent, reliable, and applicable across different clinical environments \citep{Richesson2011, Bellary2014}. \citet{Rinaldi2025} outlines essential guidelines for CRF design, emphasizing the need to use clear, reusable, standardized, and uniquely identifiable terms to facilitate semantic consistency and future reuse. In a related line of work, \citet{LIN201549} proposes methods to ensure that CRFs are aligned with the specific research questions they aim to address, thereby reinforcing their utility and validity in clinical studies.\\
The shift from paper-based to electronic CRFs has been a major focus of recent research, aiming to enhance usability, reduce errors, and improve integration with digital health records \citep{Fleischmann2017}. This advancements lead to a gain of interest about automatic CRF filling from clinical reports. \citet{MacKenzie2016} introduced early approaches to extract structured data from narrative clinical notes, a line of research that has been extended by \citet{Gutierrez-Sacristan2024}.
However, these approaches remain relatively basic, depending on keyword matching and vocabulary-based resolution, failing to leverage the full capabilities of modern Natural Language Processing techniques.
 
\section{Methodology} \label{sec:methodology}
In this section, we present a general methodology to convert corpora of annotated clinical cases into structured Case Report Forms.
Our approach is informed by an analysis of 200 pairs of clinical notes and populated CRFs from an Italian hospital. The CRFs at hand were organized among seven key areas: patient history, clinical examination, diagnostic tests results, laboratory test results, imaging findings, treatment, and final diagnosis. While CRFs are designed to be broad and comprehensive, covering a wide range of potential clinical scenarios, an individual patient’s history is typically much more limited. For this reason, we observed that in our sample the CRF items remained unfilled around 90\% of the time when populated with patients' information, highlighting the general characteristic of being designed to collect much more information of what it is typically available for each specific patient.\\
From this analysis, we concluded that in our setup CRF design lies between two extremes: creating a unique CRF for each clinical case, leading to highly specific yet non-generalizable item sets, or crafting a single, overly broad CRF for the entire dataset, potentially blending unrelated medical domains.
We adopted an intermediate approach, aligning with the traditional purpose of CRFs in clinical studies — to gather data from patients with similar conditions relevant to a study \citep{Bellary2014}. 
Building on this principle, we propose a two step procedure as outlined in Figure~\ref{fig:pipeline}: in Section~\ref{sec:groups} we group clinical cases based on semantic similarity, and in \ref{sec:crf_generation} we generate a dedicated CRF for each group and fill it with the information annotated for each clinical note. This results in one set of CRF items per group, subsequently filled once for each clinical case in that group. 
To conclude, in Section~\ref{sec:models} we introduce and detail the task, the evaluation metrics, and the method provided as baselines.\\

\subsection{Clinical Cases Clustering} \label{sec:groups}

We aim to generate groups of clinical cases, ensuring both clinical relevance and consistency in the resulting crafted CRFs.
Therefore, we require effective differentiation of documents to form clusters that group together only relevant clinical cases. If the clusters are too broad, meaningful distinctions may be lost. 
We prioritized diagnosis as the key clustering dimension since CRF items are typically guided by the specific condition being studied.
The key idea is to give significant weight to diagnosis-based links between notes in the clustering process, while retaining knowledge about entities and clinical information. Grouping documents that share similarities in these aspects helps construct synthetic CRFs that are both structured and clinically relevant.\\
Since many available datasets do not include explicit annotations on diagnoses, we implemented an automated system to extract them.

\paragraph{Diagnosis extraction.}
Extracting a diagnosis from a clinical note presents several challenges. Firstly, a note may mention past diagnoses that are no longer relevant. Secondly, the diagnosis might be implied rather than explicitly stated, requiring a deeper interpretation. Lastly, some clinical notes may not include a diagnosis at all, further complicating the extraction process.
To address this challenge, we implement a two-step approach:  
\textit{i) Automatic Generation of a Shortlist of Potential Diagnoses} – We leverage the available annotations to identify candidate diagnoses for each clinical case. First, we extract all words with the prefix "diagnos-” and check whether they are followed by an annotated entity. When this pattern was present,
the associated entity is considered a potential diagnosis. Otherwise, we treat all entities in the text as potential diagnoses.  
\textit{ ii) Diagnosis Selection} – We refine the diagnosis by prompting a Large Language Model with the shortlist. This step outputs the exact diagnosis from the shortlist, combining the pattern-matching findings and powerful models, improving accuracy and reducing ambiguity.

\begin{figure}[t]
\centering
\includegraphics[width=1\linewidth]{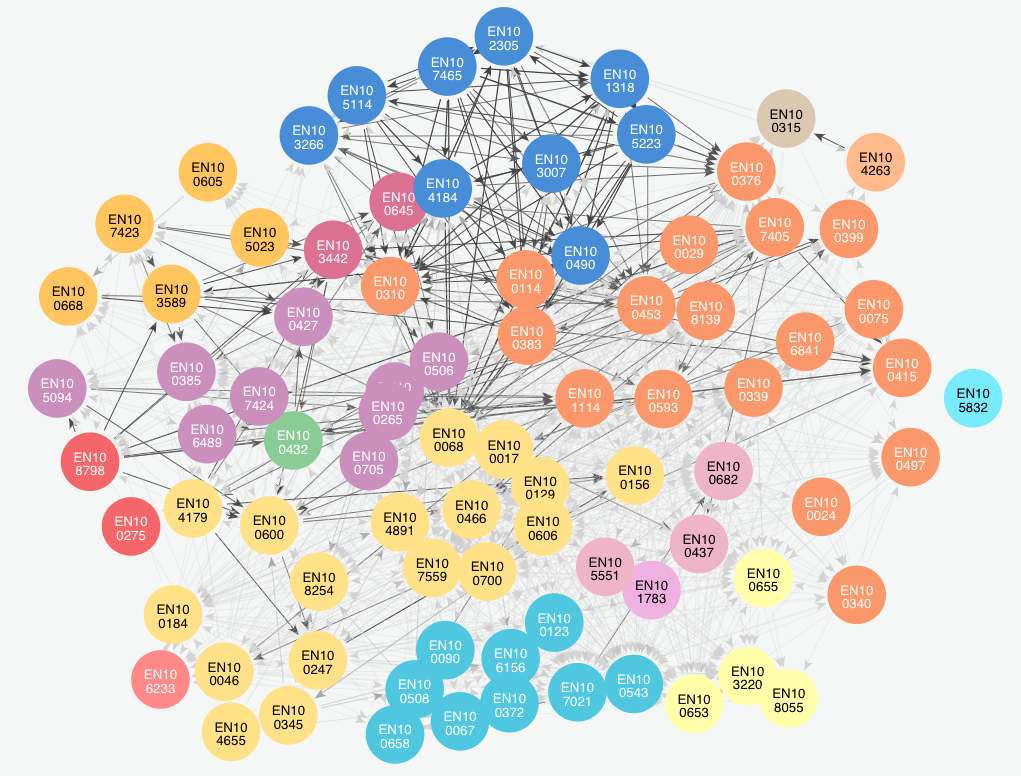}
\caption{Graph representation of the E3C English dataset. Each node is a clinical note and the edges represent the similarity between cases. Darker edges represent higher similarity. The color of the nodes represents the group assigned by the clustering algorithm.
}
\label{fig:graph}       
\end{figure}

\paragraph{Data representation for clustering.} Our clustering approach is built on a graph-based representation of the data, where clinical notes are linked by weighted edges that quantify their similarity (see Figure~\ref{fig:graph} for an implementation example of such concept). This similarity is calculated based on shared entities and diagnoses across cases. A key challenge lies in the variability of how these concepts are mentioned, as the same notion can be expressed in multiple ways (e.g., “lower limb” vs. “leg", “malignant tumor” vs. “cancer"). Ensuring that notes discussing the same or closely related concepts achieve high similarity beyond mere character overlap is a critical aspect of our methodology.
To address this challenge, we leveraged the UMLS Metathesaurus Names database \citep{UMLS2024}, augmenting the terms with semantically related concepts. 
By mean of appending to each term a short list of related ones (maximum $5$), we can better capture the similarities between cases, even when different terminology is used to refer to the same or closely related clinical concepts.
For languages other than English, each target mention is translated into English before performing a semantic search using a state-of-the-art language model \cite{zhang2025jasperstelladistillationsota} following the findings of \citet{CHIARAMELLO201622}.



\paragraph{Similarity definition.} To create the connection between each pair of clinical cases, we determine a similarity measure based on two components: the ratio of shared entities ($e$), and diagnosis similarity ($d$). The ratio $e$ is calculated as the number of UMLS-augmented shared terms divided by the number of augmented terms in the clinical note with the least of them. However, assessing diagnosis similarity $d$ requires a different strategy due to the limited number of diagnosis terms per note. We address this using a large language model trained for semantic similarity \cite{lee2024nvembed}, calculating cosine similarity between the UMLS-augmented diagnosis embeddings. This approach enables us to establish meaningful connections between cases, forming more coherent clusters. \\
We then define the overall similarity measure
\begin{equation}
s = 3 d + e
\end{equation}
This formulation assigns greater weight to diagnosis similarity while still preserving additional contextual information on shared entities.\\

\paragraph{Clustering.} Based on the overall similarities $s$, we propose to apply the Louvain algorithm as described by \citet{LU201519}, selecting as starting groups the ones composed by the weakly connected sub-graphs obtained via the $d$ edges with high weight. However, this step is highly data-dependent and must be tailored to each specific use case, following the approaches described by \citet{Xu2015}.

\subsection{CRF generation} \label{sec:crf_generation}

For each group of clinical notes, we aim to extract a set of relevant items for each section identified in the real-world CRFs analyzed in Section~\ref{sec:methodology}. The combination of the distinct section sets forms a comprehensive, group-specific CRF, tailored to the shared characteristics and clinical context of each group. Once each group-specific CRF is created, it needs to be populated for each clinical case. The overall outcome of this stage is one CRF per group and one gold-standard filled CRF per clinical case. Clinical cases within the same group share the same set of items, but their values vary based on the specific annotations present in each document.\\
We formulate and populate items for the identified sections, acknowledging that not all sections may be available in every dataset. As such, it is essential to determine which sections can be populated on the basis of the available annotations and, when necessary, refine the process to suit specific use cases. Here, we present an overview of the possible scenarios.
\textit{Clinical history} items can be generated using annotations such as symptom, sign, clinical entity, disease, condition, procedure. They are typically filled with positive and negative values, based on whether they occurred in the patient's past. Additionally, they may include information on whether a disease or condition is chronic or acute. 
\textit{Clinical examination},\textit{ diagnostic test results},\textit{ laboratory test results}, and\textit{ imaging findings} can be addressed using any annotation of type similar to condition, measurement. Such items can be populated with diverse answer formats, including numerical values, categorical labels (e.g., positive/negative, high/low), and free-text descriptions, depending on the nature of the test and the information available.
\textit{Diagnosis} items can be generated based on the extraction procedure described in Section~\ref{sec:groups}. This category of items is filled with either a positive or negative value.
\textit{Treatment} items can be addressed via labels such as medication, drug, or chemical. They can be filled with a variety of formats, spanning from medication names to time and duration information.\\
Initially, item sets are generated individually for each clinical note. These sets are then combined with those from other notes within the same group, forming a comprehensive and representative list of items for the entire group. Then, generated group-specific CRFs are populated for each clinical case in the group, based on the annotation, resulting in the gold-standard filled CRF.

\paragraph{Data revision.}
All generated items in each section of each group-specific CRF are normalized using UMLS mapping, collapsing equivalent terms to a single one. Furthermore, manual revision is performed to guarantee the quality of the generated CRF, with three objectives: (1) merge equivalent and highly related items, (2) remove irrelevant items, and (3) adjust inaccurate items. The process is conducted in a semi-automated manner. For each item in the CRF, we use a close source Large Language Model to assess whether it could be mapped to an existing item and to provide a justification for the suggested mapping. Any proposed mapping is manually reviewed for validity and, if approved, the overlapping items are consolidated.

\section{CRF Filling: Task Definition} \label{sec:task_def}

Datasets constructed according to the methodology detailed in Section~\ref{sec:groups} introduce a new CRF-filling task, which is divided into as many sub-tasks as the number of corresponding CRF sections. Each task requires filling CRF items based on information extracted from clinical cases, but they may vary in complexity. For the \textit{diagnosis} section, the task consists in determining whether an item represents the final diagnosis, with three possible outcomes — “yes", “no", “not available". The \textit{clinical history} section is more complex than the diagnosis one, as it allows for a broader range of valid outcomes. In addition to determining whether an event occurred in the patient's history, it may also capture details such as its chronic or acute nature, adding an extra layer of difficulty. The \textit{clinical examination}, \textit{tests results}, \textit{imaging findings} and \textit{treatment} section are the most complex ones, as they lack a predefined set of valid answers, requiring extraction and interpretation of numerical and textual values from the clinical notes.

\paragraph{Baseline.}
We established a baseline for the CRF slot filling tasks using sequence and pattern matching techniques. For the diagnosis task, the baseline assigns “yes” if the diagnosis of interest appears in the clinical case and “not available” otherwise. In the clinical examination, tests results, imaging findings and treatment tasks, if the respective item is mentioned in the text, the first numerical value following it is extracted as the result. For the clinical history task, the baseline assigns “yes” if the corresponding textual span is found in the clinical case and “not available” otherwise.

\paragraph{Metrics definition.}

To define the evaluation metrics, we first established criteria for identifying positive and negative occurrences. An item is considered positive for diagnosis if labeled “yes” and negative if marked as “no” or “not available". In the clinical history section, an item is positive if it appears in any valid form, such as “yes”  or “no", and negative if marked as “not available".  For all the other tasks, an item is positive if assigned a value and negative if labeled as “not available.” 
Additionally, when a generated answer does not conform to the expected format—if any predefined format is required—it is always considered a false positive.
Based on these definitions, we can compute task-specific precision, recall, and $F_1$-score, as well as overall micro and macro $F_1$-scores.\\
We apply strict matching criteria (ignoring trailing punctuation) with one relaxation: any text following “not available”  in response was ignored if this phrase appeared at the beginning. For gold-standard labels filled with multiple values, a true positive (TP) is assigned for a perfect match, a false positive (FP) if extra elements are predicted, and a false negative (FN) if the prediction contains fewer items than the ground truth.\\


\section{The Case of the E3C Dataset}
\label{Sec:e3c-original}

In the previous sections, we outlined the general methodology for converting any corpus annotated for information extraction into gold-standard filled CRFs. In this section, we apply this methodology to the European Clinical Case Corpus (E3C, \citealp{Magnini2023}). E3C is an open, manually annotated multilingual dataset consisting of clinical cases in five languages.
E3C clinical cases are detailed accounts of a patient's medical history, containing rich medical details and temporal relationships that enable in-depth linguistic analysis.
The dataset includes annotations on both textual spans and the relationships between them. The ones relevant to our study are summarized in Table~\ref{tab:e3c_annotation_text}. 
In this work, we focused on the Italian and English splits (Table~\ref{tab:e3c_descr}).

\begin{table}
  \centering
  \begin{tabular}{lllll}
    \hline
    \textbf{Lang} & \textbf{\# notes} & \textbf{\# clent}  & \textbf{\# rml} & \textbf{\# event}  \\
    \hline
    English    &  84  &  1024   & 480 &4885 \\
    Italian    & 86    &  869  & 383 & 3385\\
    \hline
  \end{tabular}
  \caption{Number of clinical notes (\# notes), annotated clinical entities (\# clent), results and measurements (\#rml), and events (\# event) in E3C Italian and English splits, which both comprise approximately $25$k words.}
  \label{tab:e3c_descr}
\end{table}

\begin{table*}
  \centering
\begin{tabular}{cp{10cm}}
    \hline
    \textbf{Category} & \textbf{Description (example)} \\ 
    \hline
    Clinical entity & disorders, pathologies, and symptoms (“metastases” “nausea") \\ \cmidrule(lr){1-2}
    Body part & parts of the human body (“parotid gland") \\ \cmidrule(lr){1-2}
    RML & results and measurements  (“38g/dl") \\ \cmidrule(lr){1-2}
    Event & any event (“diagnosed", “haemoglobin") \\ \cmidrule(lr){1-2}
    PERTAINS-TO & relation between an RML and the Event it refers to (“38g/dl” pertains-to “haemoglobin") \\
    \hline
  \end{tabular}
  \caption{E3C categories for annotations on textual spans and their relationships utilized in this work. Each textual span is annotated if it represents a clinical term (i.e., clinical entities such as pathologies and symptoms, body parts, laboratory tests and results) and is assigned some attributes. For more details, see \citet{Magnini2023}.}
  \label{tab:e3c_annotation_text}
\end{table*}

\subsection{CRF generation from E3C} 

We applied our methodology to the European Clinical Case Corpus (E3C), adapting it to the dataset’s specific characteristics. Below, we outline key adaptations, while all details not explicitly mentioned can be found in Section~\ref{sec:methodology}.\\
We generated the shortlist of potential diagnoses considering only clinical entities as possible targets, as other annotations were deemed out of scope. After selecting the diagnoses using GPT-4o \citep{openai2024gpt4technicalreport} in a 4-shots settings, we manually reviewed $10$ examples in both English and Italian, confirming the accuracy of the results in all cases. In some instances ($9$ for English, $19$ for Italian), no diagnosis was identified, which is expected since certain clinical documents do not report it.
Then, the overall similarity measure was defined as $s = 3 d + \frac{1}{2}(e+b)$, where $e$ and $b$ are the ratios of shared clinical entities and shared body parts respectively, $d$ is the diagnosis similarity.
The resulting graph representation of the data is shown in Figure~\ref{fig:graph}.
This method resulted in 7 (8) groups and 6 (12) clinical cases not assigned to any group for Italian (English). More details on the diagnosis extraction prompts, similarities and generated groups are shown in Appendix~\ref{app:diag_extr} and ~\ref{app:ex_crf}.\\
Using the information embedded in the E3C annotations, we formulated and populated items for the following sections:
clinical history, diagnosis, clinical examination, diagnostic test results, laboratory test results, and imaging findings. Since no information on treatment was available at the annotation level, we excluded it from consideration.
\paragraph{Exams.} To generate and populate exam items, we first extracted the textual spans linked to RMLs (results and measurements) via PERTAINS\_TO relationships. A CRF exam item was created for each textual span with a corresponding RML, representing its filling value.
When an RML refers to multiple textual spans, a separate item is generated for each of them. When the same textual span is associated with multiple RMLs, a single item is created for the textual span, and each RMLs is used at filling time, separated by special tokens.
RMLs that do not pertain to any textual span were ignored.

\paragraph{Clinical History.} To generate and populate items about patients history, we focused on the clinical entities enriched by three key annotated attributes: “polarity” (whether the reported term is present or not), “contextual modality” (knowledge about the truth value of the event, can be actual, hedged, hypothetical or generic), and “permanence” (can be permanent for conditions with no known cure or finite for those that can be resolved eventually). Each of these attributes defines a portion of the gold-standard answer, as outlined in Table~\ref{tab:attributes_hist} in Appendix~\ref{app:hist}.
\paragraph{Diagnosis.}
For each diagnosis, an item was created and populated with “yes” if it applied to the clinical case and “not available” otherwise.\\
An example of a generated CRF can be found in Appendix~\ref{app:ex_crf}.

\begin{table*}
  \centering
  \begin{tabular}{l|p{3.8cm}|p{3.6cm}|p{1cm}p{1cm}|p{1cm}p{1cm}}
    \hline
     & & & \multicolumn{2}{c|}{\textbf{Italian}} & \multicolumn{2}{c}{\textbf{English}} \\
    \hline
      & & & \multicolumn{1}{c}{Train} & \multicolumn{1}{c|}{Test} & \multicolumn{1}{c}{Train} & \multicolumn{1}{c}{Test} \\
    \textbf{Task} & \textbf{Description}& \textbf{Accepted answers} & \textbf{Items (Filled)} & \textbf{Items (Filled)} &  \textbf{Items (Filled)}& \textbf{Items (Filled)}  \\
    \hline
    Diagnosis & determine if an item is the final diagnosis for the patient  & “yes", “no", “not available” & 498 (8\%)    & 553 (7\%)    & 491 (9\%)   & 505 (9\%)   \\
    \hline
    History & determine whether the patient experienced a history item   & “Certainly yes", “No", "Probably yes, chronic", “not available” etc.& 977  (23\%)   & 903    (11\%)   & 953    (25\%)     & 872   (13\%)  \\
    \hline
    Exams   & extract the results related to an exam item & any string representing an exam result & 1108   (10\%)  & 1149   (10\%)   & 984    (11\%)  & 916    (9\% )  \\
    \hline
    \textbf{Total} & &  & \textbf{2583}   (\textbf{14\%})  & \textbf{2605}   (\textbf{10\%})  & \textbf{2428}   (\textbf{16\%})  & \textbf{2293}   (\textbf{11\%})  \\
    \hline
  \end{tabular}
  \caption{Description, space of possible answers, number of items, and ratio of populated items in the train and test splits for both languages for the three CRF sub-tasks. All three sub-tasks are quite sparse, with around ten to fifteen percent of the items populated in the gold-standard filled CRFs. Clinical notes in the train and test split are composed by around 12k and 13k tokens (words), respectively, in both Italian and English. The possible answers for history are determined by the levels of the annotated attributes utilized for the gold-standard filling.}
  \label{tab:resulting_dataset}
\end{table*}

\paragraph{Train-test split.}
We adopted the train-test split provided by \citet{Ghosh2025} for the E3C dataset. The result is that clinical cases from the same group are assigned to different splits, while group-specific CRFs are generated on all the cases in the corpus.
By design, CRFs must cover all essential fields for the patient groups they represent. As a result, constructing comprehensive item sets from the full dataset is necessary and does not introduce bias beyond the task's inherent structure. Crucially, only training clinical notes are used for learning, preventing any test-specific influence on the model.
Note that this cross-splits effect is further reduced by creating clinical history item sets merging the ones extracted from clinical cases in both splits but excluding from the final set the ones filled only for test cases after data revision.  

\begin{table*}
  \centering
  \begin{tabular}{l|ccc|ccc|ccc|cc}
    \hline
    \textbf{Model} & \multicolumn{3}{c|}{\textbf{Diagnosis}} & \multicolumn{3}{c|}{\textbf{History}} & \multicolumn{3}{c|}{\textbf{Exams}} & \textbf{Micro} & \textbf{Macro} \\
    \cline{2-4} \cline{5-7} \cline{8-10}
    & Prec. & Recall & $F_1$ & Prec. & Recall & $F_1$ & Prec. & Recall & $F_1$ & \textbf{$F_1$} & \textbf{$F_1$} \\
    \hline
    Baseline & 64.9 & 58.5 & 61.5 & 100.0 & 11.3 & 20.4 & 13.6 & 10.8 & 12.0 & 31.3 & 25.4 \\
    Llama 8B & 32.2 & 92.7 & 47.8 & 7.2 & 60.8 & 13.0 & 4.8 & 25.0 & 8.1 & 23.0 & 18.2 \\
    Qwen 7B & 72.1 & 75.6 & 73.8 & 33.8 & 73.6 & 46.4 & 7.5 & 8.5 & 7.9 & 42.7 & 35.2 \\
    Mistral 24B & 68.4 & 63.4 & 65.8 & 51.6 & 64.9 & 57.5 & 13.8 & 22.1 & 16.9 & 46.7 & 41.4 \\
    Gemma 27B & 73.5 & 87.8 & \textbf{80.0} & 47.1 & 83.5 & \underline{60.2} & 22.9 & 83.9 & \underline{36.0} & \underline{58.7} & \underline{53.7} \\
    Llama 70B & 54.7 & 100.0 & 70.7 & 32.8 & 77.3 & 46.0 & 16.0 & 67.8 & 25.9 & 47.5 & 42.4 \\
    Qwen 72B & 75.6 & 75.6 & 75.6 & 58.1 & 74.2 & \textbf{65.2} & 19.4 & 38.7 & 25.8 & 55.5 & 50.0 \\
    GPT 4o & 75.6 & 82.9 & \underline{79.1} & 40.8 & 75.3 & 52.9 & 34.0 & 76.8 & \textbf{47.1} & \textbf{59.7} & \textbf{55.9} \\
    \hline
  \end{tabular}
  \caption{Performance of different models on the Italian dataset across three categories: Diagnosis, History, and RML. Metrics include Precision, Recall, $F_1$-score. Overall micro and macro $F_1$-scores are also reported.}
  \label{tab:model_performance_ita}
\end{table*}

\begin{table*}
  \centering
  \begin{tabular}{l|ccc|ccc|ccc|cc}
    \hline
    \textbf{Model} & \multicolumn{3}{c|}{\textbf{Diagnosis}} & \multicolumn{3}{c|}{\textbf{History}} & \multicolumn{3}{c|}{\textbf{Exams}} & \textbf{Micro} & \textbf{Macro} \\
    \cline{2-4} \cline{5-7} \cline{8-10}
    & Prec. & Recall & $F_1$ & Prec. & Recall & $F_1$ & Prec. & Recall & $F_1$ & \textbf{$F_1$} & \textbf{$F_1$} \\
    \hline
    Baseline & 84.6 & 53.7 & 65.7 & 87.5 & 13.2 & 23.0 & 0.0 & 0.0 & 0.0 & 29.6 & 21.9 \\
    Llama 8B & 49.3 & 94.9 & 64.9 & 10.2 & 76.4 & 18.0 & 6.4 & 63.0 & 11.6 & 31.5 & 25.1 \\
    Qwen 7B & 100.0 & 63.4 & 77.6 & 40.0 & 78.4 & 53.0 & 15.6 & 16.3 & 15.9 & 48.8 & 41.9 \\
    Mistral 24B & 63.6 & 80.0 & 70.9 & 55.3 & 68.9 & \textbf{61.3} & 22.7 & 62.5 & 33.3 & 55.2 & 51.0 \\
    Gemma 27B & 91.4 & 78.0 & \underline{84.2} & 42.9 & 74.5 & 54.5 & 32.7 & 86.0 & 47.4 & \underline{62.0} & \underline{57.7} \\
    Llama 70B & 84.2 & 78.0 & 81.0 & 36.1 & 74.3 & 48.6 & 34.8 & 81.6 & \underline{48.8} & 59.5 & 55.6 \\
    Qwen 72B & 96.8 & 73.2 & 83.3 & 55.9 & 67.0 & \underline{60.9} & 27.0 & 80.0 & 40.3 & 61.5 & 56.6 \\
    GPT 4o & 94.4 & 82.9 & \textbf{88.3} & 47.5 & 72.4 & 57.4 & 42.2 & 84.3 & \textbf{56.2} & \textbf{67.3} & \textbf{63.4} \\
    \hline
  \end{tabular}
  \caption{Performance of different models on the English dataset across three categories: Diagnosis, History, and RML. Metrics include Precision, Recall, and $F_1$-score. Overall micro and macro $F_1$-scores are also reported.}
  \label{tab:model_performance_eng}
\end{table*}

\section{Experimental settings}

We explored the E3C CRF-filling task using decoder-only Large Language Models (LLMs) as they have exhibited high performance in several tasks in zero-shot settings. 

\paragraph{Models.}
We selected the instruct versions of different state-of-the-art model families, in different sizes: Llama-3 8B and 70B \citep{grattafiori2024llama3herdmodels}, Qwen-2.5 7B and 72B \citep{qwen2025qwen25technicalreport},  Mistral-Small-3.1 24B \footnote{\url{https://mistral.ai/news/mistral-small-3-1}}, Gemma-3 27B\footnote{\url{https://blog.google/technology/developers/gemma-3/}} and GPT 4o. This selection allowed us to compare proprietary (GPT) and open-source models (the others), assessing the impact of model size and determining which family performs better on our task. Each model was prompted with task-specific details, the clinical case, the CRF item, and answering guidelines. \\

All experiments on open-source models were run on 8xA40 (46GB) and took approximately $30$ GPU hours, serving the models using the vllm package \citep{kwon2023efficient}.
Prompts can be seen in detail in Appendix~\ref{app:prompt}.

\subsection{CRF Filling from E3C Clinical Cases} \label{sec:models}

The constructed dataset introduces a new E3C CRF-filling task, which is divided into three sub-tasks: clinical history, exams, and diagnosis as described in Table~\ref{tab:resulting_dataset}. 
The main specialty of this task in respect to the more general outlined in Section~\ref{sec:task_def} is that clinical history items can be filled with twelve valid values (Appendix~\ref{app:hist}).
Given the unique annotation scheme in E3C, which includes multiple levels of polarity, contextual modality, and permanence, such complexity is specific to this dataset and may not be present in others. Therefore, we report results on a simplified version where all positive responses are grouped as “yes” and all negative ones as “no". By simplifying the values, we aim to offer a more general perspective on the inherent difficulty of the task, extending beyond the particularities of the E3C dataset.

%
%

\section{Results and Discussion} \label{sec:discussion}

Experimental results are reported in Tables~\ref{tab:model_performance_ita} and ~\ref{tab:model_performance_eng} for Italian and English respectively.
GPT-4o consistently achieves the highest overall performance in both Italian and English datasets, with the best Micro and Macro $F_1$-scores (59.7 and 55.9 for Italian, 67.3 and 63.4 for English). 
Among open-weight models, Gemma 27B and Qwen 72B perform competitively in Italian, closely approaching GPT-4o’s results, particularly in diagnosis and history. 
For English, Gemma 27B, Qwen 72B, and Llama 70B performances are very similar, around 6-8 points lower than GPT-4o.\\
Regarding model size, we observe an average improvement of around 20 Macro $F_1$ points when scaling from small (7/8B) to large (70/72B) models in the LLaMA and Qwen families. Interestingly, models in the 20–30B range often match or surpass larger ones from different architectures. Among the tasks, exams prove to be the most challenging, followed by history, indicating significant room for improvement. Models perform on average better on English than in Italian with no exception, with an average delta of 7.5 points of Micro $F_1$.
Among the smaller models, Qwen 7B significantly outperforms Llama 8B, which struggles with extremely low precision. At larger scales, Qwen 72B and Llama 70B exhibit comparable performance in English, while Qwen 72B demonstrates a clear advantage over Llama 70B in Italian.

\section{Conclusion}
Our study presents a novel methodology for transforming annotated clinical notes into structured Case Report Forms (CRFs) by leveraging clusters of semantically similar cases. This approach ensures that CRFs are both comprehensive and contextually relevant while maintaining consistency across similar clinical scenarios. Given the scarcity of publicly available CRF datasets, our method provides a valuable framework for automating CRF generation, which could be highly beneficial for future clinical applications.
In addition, our method brings existing datasets closer to real-world clinical applications, ensuring greater alignment with the practical needs of hospitals and research. Given that CRF filling is a widely relevant task, this approach enhances the utility of annotated clinical notes.

Our findings highlight that the characteristics of the generated CRFs are strongly influenced by the dataset's distribution, underscoring the necessity of manual tuning based on available annotation types when adapting the method to different contexts.
We believe that a robust analysis of the data distribution is crucial for high-quality CRF generation.

Our experimental results reveal that the constructed CRFs encompass tasks of increasing complexity for state-of-the-art models. Diagnosis items can be framed as a relatively straightforward binary classification task, while history items remain within a classification framework but with greater difficulty due to their nuanced nature. The most challenging aspect lies in handling exams, tests, and examinations, which require a fully generative approach without a predefined set of valid responses, making them particularly difficult for current models to solve. Both open- and closed-source models show room for improvement in terms of performance.

\section*{Limitations}
There are a few limitations in our current approach to convert Information Extraction datasets into structured CRFs. First, the proposed methodology has been experimented only on the E3C corpus: although this is a significant use case (several levels of annotations, several languages), additional insights may derive from different available datasets. Second, in order to keep under control our experimental setting, we made a few simplifications with respect to the full complexity of the task. Particularly, for the CRF \textit{clinical history} group, we assumed a three-value schema (i.e., a certain clinical evidence is either present, negated, or not mentioned), while in reality the possible values should be extended to cover cases of chronicity.

\section*{Acknowledgments}
This work has been partially funded by the European Union under the Horizon Europe eCREAM Project (Grant Agreement No.101057726) and IDEA4RC Project (Grant Agreement No.101057048). Views and opinions expressed are however those of the authors only and do not necessarily reflect those of the European Union or the European Health and Digital Executive Agency (HADEA). Neither the European Union nor the granting authority can be held responsible for them.

\bibliography{anthology,custom, crf}

\appendix

\section{Appendix}
\label{sec:appendix}

\subsection{Diagnosis Extraction}\label{app:diag_extr}

Here we report the structure of the prompt utilized to generate the diagnosis using GPT-4o: 
\begin{verbatim}
{System prompt}{Example 1}...{Example 4}
"clinical note":{Clinical case}
"potential diagnosis":
{list of potential diagnosis}
\end{verbatim}

Here is the system prompt:
\begin{verbatim}
You are a clinical assistant. 
Your job is to extract the conclusive 
diagnosis from a clinical note written by 
an experienced physician. 
The diagnosis is a medical condition 
identified by a health care provider. 
To complete the task, you are aided by a 
list of possible diagnoses. 
Here are your guidelines:
1. The diagnosis is always contained in 
the list of potential diagnoses.
2. Your goal is to extract only the 
diagnosis, ignoring everything else. 
3. Respond with a json containing the 
extracted diagnosis and a short motivation
{“motivation”: “motivation for the 
extracted diagnosis”, “diagnosis”: 
“extracted diagnosis”}.
4. If no diagnosis is reported, 
respond with “no diagnosis.” 

CAUTION: Notes may contain diagnoses 
made in the past with respect to 
the current clinical situation. 
Only extract diagnoses related to the current 
situation.
\end{verbatim}

\noindent Table~\ref{tab:similarity_scores} presents examples of similarity scores for E3C cases calculated in the embedding space of the diagnosis augmented via UMLS semantic search.

\subsection{E3C Clinical History Items}  \label{app:hist}

Table~\ref{tab:attributes_hist} reports the attributes and their levels used for populating the E3C CRF clinical history section. Each E3C clinical entity is annotated with contextual modality, polarity, and permanence, which determine the filled value using the template: 
\begin{verbatim}
{contextual mod} {polarity}, {permanence}
\end{verbatim}
For instance, an entity with polarity ``positive", contextual modality ``hedged" and permanence ``finite" is filled with ``Probably yes, possibly chronic". There are $12$ possible level combinations.

\begin{table}
  \centering
  \begin{tabular}{l|l|p{2cm}}
    \hline
    \textbf{Attribute} & \textbf{Level} & \textbf{CRF Value} \\
    \hline
    \multirow{2}{*}{Polarity} & Positive & Yes \\
                               & Negative & No \\
    \hline
    \multirow{4}{*}{Modality} & Actual & Certainly \\
                                          & Hypothetical & Possibly \\
                                          & Hedged & Probably \\
                                          & Missing & (empty) \\
    \hline
    \multirow{3}{*}{Permanence} & Permanent & Chronic \\
                                 & Finite & Certainly not chronic \\
                                 & Missing & Possibly chronic \\
    \hline
  \end{tabular}
  \caption{Attribute levels for populating the E3C CRF clinical history section.}
  \label{tab:attributes_hist}
\end{table}

\begin{table*}[h]
    \centering
    \begin{tabular}{p{5cm}|p{5cm}|l}
        \hline
        \textbf{Diagnoses note 1} & \textbf{Diagnoses note 2} & \textbf{Similarity Score} \\
        \hline
        neuroendocrine neoplasia & neoplasia & 0.63 \\ \hline
        chronic myeloid leukemia Ph+ in chronic phase & JMML & 0.57 \\   \hline
        acute ulcerative rectocolitis & clostridium difficile colitis & 0.58 \\   \hline
        mass of tumor origin & syncopal episodes, Polymorphic ventricular tachycardia & 0.11 \\   \hline
        Wilms's tumor, Metastasis & microperforation & 0.10 \\
        \hline
    \end{tabular}
    \caption{Similarity scores between extracted diagnoses for pairs of clinical cases. The first three lines represent cases with high similarity, while the last two cases with low similarities. It can be noted that terms that are syntactically different but semantically close such as  “JMML” and “Chronic myeloid leukemia Ph+ in chronic phase” are mapped together by this approach, as the former has been correctly enriched with the term “juvenile myeloid leukemia", that results in an embedding similar to the latter. At the same time, cases with very different diagnoses are assigned very low similarities.}
    \label{tab:similarity_scores}
\end{table*}


\subsection{Generated E3C CRFs} \label{app:ex_crf}
\begin{figure*}[t]
\center
  \includegraphics[width=0.9\linewidth]{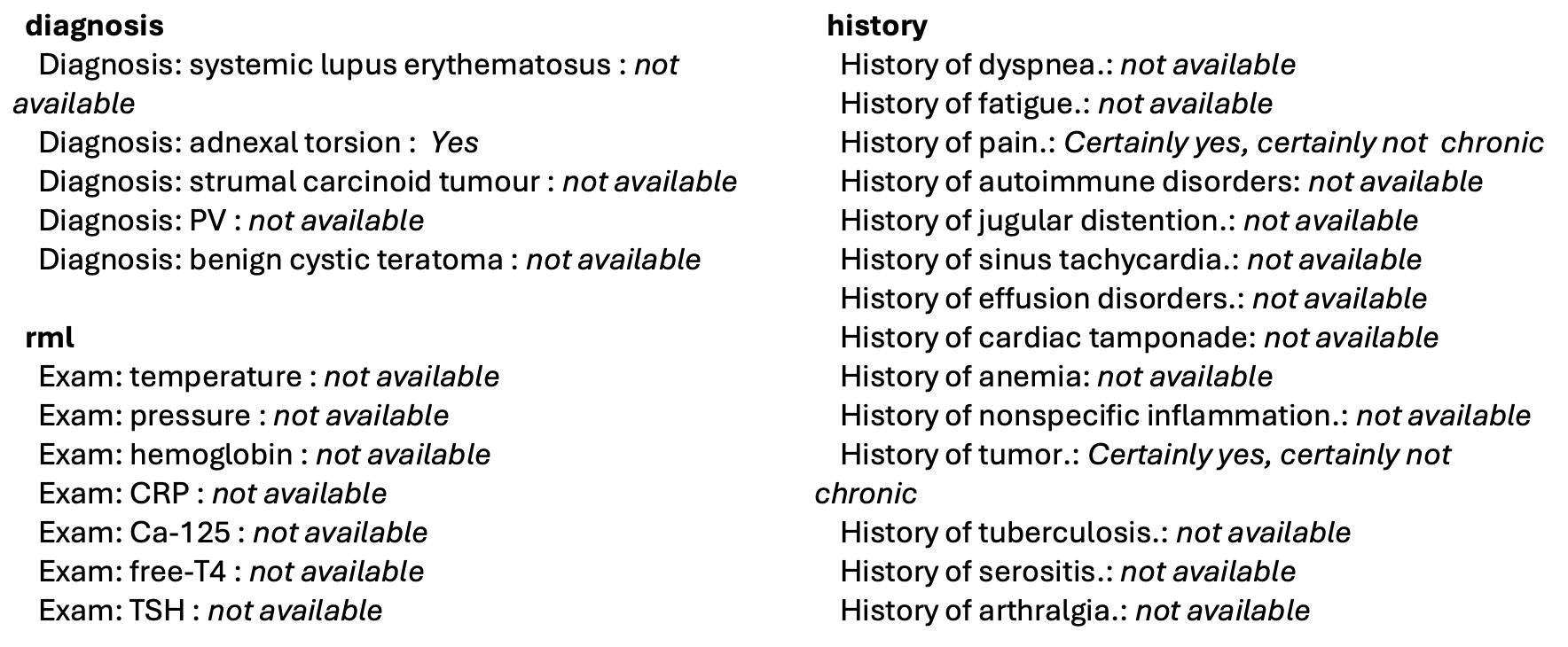}
  \caption{Example of a generated CRF for English group 1 and filled with the annotation from the clinical case EN100668}
\label{fig:crf_ex}       
\end{figure*}

Table~\ref{tab:crf_by_group} presents the statistics on the generated E3C CRF for English and Italian.
Figure~\ref{fig:crf_ex} shows an example of a CRF generated for the English group 1.

\begin{table*}[h]
    \centering
    \begin{tabular}{p{0.8cm}p{1.3cm}p{0.8cm}p{1.5cm}p{1.5cm}|p{0.8cm}p{1.3cm}p{0.8cm}p{1.5cm}p{1.5cm}}
        \hline
        \multicolumn{5}{c|}{\textbf{Italian}} & \multicolumn{5}{c}{\textbf{English}} \\
        \hline
        \textbf{Group} & \textbf{Cases Train/Test}& \textbf{CRF items} & \textbf{Avg/StDev (Train)} & \textbf{Avg/StDev (Test)} & \textbf{Group} & \textbf{Cases  Train/Test} & \textbf{CRF items} & \textbf{Avg/StDev (Train)} & \textbf{Avg/StDev (Test)} \\
        \hline
        0 &   4/4 & 23 & 5.5 / 2.2 & 5.5 / 0.8 & 0  & 7/2 & 71 & 11.7 / 4.7 & 5.0 / 3.0 \\
        1 & 11/13 & 91 & 7.9 / 5.3 & 7.9 / 3.6 & 1  & 1/4 & 26 & 19.0 / 0.0 & 3.0 / 1.2 \\
        2 &   4/4 & 55 & 13/ 9.7 & 13 / 2.2 & 2 & 1/1 & 10 & 6.0 / 0.0 & 4.0 / 0.0 \\
        3 &   2/7 & 27 & 4.5 / 1.5 & 4.5 / 2.7 & 3  & 3/6 & 54 & 11 / 9.7 & 6.8 / 5.2 \\
        4 &   4/6 & 76 & 9.8 / 4.8 & 9.8 / 9.2 & 4  & 5/4 & 24 & 3.6 / 1.5 & 4.0 / 2.5 \\
        5 &   4/4 & 48 & 9.8 / 7.8 & 9.8 / 4.5 & 5  & 9/9 & 99 & 11 / 7.7 & 7.9 / 4.0 \\
        6 &   9/4 & 79 & 12.7/ 7.5 & 13 / 1.2 & 6 & 8/9 & 36 & 9.0 / 4.0 & 7.0 / 5.1 \\
          &       &    &           &            & 7 & 2/2 & 75 & 9.4 / 5.4 & 13 / 11 \\ 
        \hline
    \end{tabular}
    \caption{Number of cases, number of items, average and standard deviation of the number of populated items (i.e., different from ``not available") per group-specific CRF for both languages.}
    \label{tab:crf_by_group}
\end{table*}

\subsection{Prompts for experiments} \label{app:prompt}

The prompt for the experiments is composed following this template: 
\begin{verbatim}
{system prompt} {answering guidelines} 
{clinical case} {question on the item}.
\end{verbatim}
Here we report the prompts used for English. The ones for Italian are the direct translation of them.

\paragraph{System prompt}
\begin{verbatim}
You are an expert clinical doctor. You have
to answer a question on "{task_description}" 
about a patient. To do it, you are given 
the patient clinical history.
\end{verbatim}

\paragraph{History answering guidelines}, where values are populated according to the logic presented in the methodology section.
\begin{verbatim}
The answer is composed by three components:
polarity, contextual modality, and 
permanence. You must combine these three 
components together to answer the question.
- contextual modality can be: 
a)'VALUE_1' if the answer is certain, 
b)'VALUE_2' if the answer is hypothetical,
c)'VALUE_3' if the answer is probable.
- polarity can be: 
a)'VALUE_4' if the answer is affirmative, 
b)'VALUE_5' if the answer is negative.
- permanence can be: 
a)'VALUE_6' if the object of the question 
is certainly permanent forever, 
b)'VALUE_7' if the object of the question 
is temporary or transitory, 
c)'VALUE_8' otherwise.

These three components must be combined 
in order: "contextual modality polarity, 
permanence". For example, if the question 
is "Does the patient have a history of 
diabetes?", the answer could be: 
"EXAMPLE_1", or "EXAMPLE_2".

If the information is not contained in 
the clinical history, answer with 
'not_available'.
Do not add any preamble to the answer.
\end{verbatim}

\paragraph{Exams answering guidelines}
\begin{verbatim}
The answer can assume three different 
formats.
-if the test/exam has been performed 
only once, answer with the results 
of the test/exam.
-if the test/exam has been performed 
more than once, report all the 
results separated by the special 
token  [\MULTI_ANSWER]  (for example 
"RESULT_1 [\MULTI_ANSWER] RESULT_2").
-if the information is not contained in 
the clinical history, answer 
with 'not_available'
\end{verbatim}

\paragraph{Diagnosis answering guidelines}

\begin{verbatim}
Answer 'Yes' if the patient's definitive 
diagnosis is the one indicated. If the 
information is not contained in the 
clinical history, answer with 'not_available'.
\end{verbatim}

\paragraph{Question structure for exams}
\begin{verbatim}
What are the results and measures of {item}?
\end{verbatim}

\paragraph{Question structure for diagnosis}
\begin{verbatim}
Is the definitive diagnosis {item}?
\end{verbatim}

\paragraph{Question structure for history}
\begin{verbatim}
Does the patient have a history 
of {item}?
\end{verbatim}

\end{document}